\documentclass[a4paper,11pt,fleqn]{article}

\usepackage[ansinew]{inputenc}
\usepackage{hyperref}
\usepackage[mathscr]{eucal}
\usepackage{graphicx}
\usepackage[font=footnotesize,labelsep=period,justification=justified,figureposition=bottom,tableposition=top]{caption}
\usepackage{subcaption}
\usepackage{multirow}
\usepackage{amsmath,amssymb,amsthm}
\usepackage{comment}
\allowdisplaybreaks

\hypersetup{colorlinks, linkcolor=blue, citecolor=blue, urlcolor=blue}
\newcommand{\todo}[1][\null]{\ensuremath{\clubsuit}}

\newcommand{\noprint}[1]{}

{\theoremstyle{definition}

\newtheorem*{remark*}{Remark}
}

\newcommand{\checked}[1][\null]{\ensuremath{\boldsymbol{\surd}}}

\begin{document}

\par\noindent {\LARGE\bf
Improving physics-informed neural networks \\
with meta-learned optimization
\par}

\vspace{4mm}\par\noindent {\large
Alex Bihlo$^\dag$
\par}

\vspace{4mm}\par\noindent{\it
$^{\dag}$Department of Mathematics and Statistics, Memorial University of Newfoundland,\\
$\phantom{^{\dag}}$~St.\ John's (NL) A1C 5S7, Canada
}

\vspace{2mm}\par\noindent {\it
\textup{E-mail:} abihlo@mun.ca
}\par

\vspace{12mm}\par\noindent\hspace*{10mm}\parbox{140mm}{\small
We show that the error achievable using physics-informed neural networks for solving differential equations can be substantially reduced when these networks are trained using meta-learned optimization methods rather than using fixed, hand-crafted optimizers as traditionally done. We choose a learnable optimization method based on a shallow multi-layer perceptron that is meta-trained for specific classes of differential equations. We illustrate meta-trained optimizers for several equations of practical relevance in mathematical physics, including the linear advection equation, Poisson's equation, the Korteweg--de Vries equation and Burgers' equation. We also illustrate that meta-learned optimizers exhibit transfer learning abilities, in that a meta-trained optimizer on one differential equation can also be successfully deployed on another differential equation.
\par}\vspace{7mm}

\section{Introduction}\label{sec:IntroductionLearnableOptimization}

Physics-informed neural networks are a class of methods for solving systems of differential equations. Originally proposed in the 1990s~\cite{laga98a} and popularized through the work~\cite{rais19a}, physics-informed neural networks have seen an immense raise in popularity in the past several years. This is in part due to the overall rise in interest in all things related to deep neural networks~\cite{lecu15a}, but also due to some practical advantages of this method compared to traditional numerical approaches such as finite difference, finite elements or finite volume methods. These advantages include the evaluation of derivatives using automatic differentiation~\cite{bayd18a}, their mesh-free nature and an overall ease of implementation through modern deep-learning frameworks such as \texttt{TensorFlow} or \texttt{PyTorch}. Given the expressive power of deep neural networks~\cite{cybe89a}, neural networks are also a well-suited class of function approximation for the solution of systems of differential equations. 

A main downside of physics-informed neural networks is that a complicated optimization problem involving a rather involved composite loss function has to be solved~\cite{rais19a}. The difficulty in solving such so-called multi-task problems is well-documented in the deep learning literature, see e.g.~\cite{yu20a}. Moreover, since essentially all methods of optimization for deep neural network today are at most of first-order, such as stochastic gradient descent, and its momentum-based flavours such as Adam~\cite{king14a}, the level of error that can typically be achieved with vanilla physics-informed neural networks as proposed in~\cite{laga98a,rais19a} is often subpar compared to their traditional counterparts used in numerical analysis. While lower numerical error can be achieved using more involved strategies, such as domain decomposition approaches~\cite{jagt20a}, modified loss functions~\cite{jin21a,wang22a} or operator-based approaches~\cite{wang23a}, all of these approaches either sacrifice some of the simplicity of vanilla physics-informed neural networks or substantially increase their training times.

Since a main culprit in of the overall unsatisfactory error levels achievable with vanilla physics-informed neural networks is the optimization method used, it is natural to aim to find better optimizers. More broadly, optimization is a topic extensively studied in the field of machine learning, with many new optimizers being proposed that aim to overcome some of the (performance or memory) shortcomings of the de-facto standard Adam, see e.g.~\cite{luca18a,shaz18a}. There has also been growing interest in the field of learnable optimization, referred to as \textit{learning to learn}~\cite{chen22a}, which aims to develop optimization methods parameterized by neural networks, that are then meta-learned on a suitably narrow class of tasks, on which they typically outperform generic (non-learnable) optimization methods.

The aim of this paper is to explore the use of learnable optimization for training physics-informed neural networks. We show that meta-trained learnable optimizers with very few parameters can substantially outperform standard optimizer in this field. Moreover, once meta-trained, these optimizers can be used to train physics-informed neural networks with minimal computational overhead compared to traditional optimizers. 

The further organization of this paper is as follows. In Section~\ref{sec:PINNsOptimization} we present a more formalized review on how neural networks can be used to solve differential equations. Section~\ref{sec:RelatedWorkOptimization} presents a short overview of the relevant previous work on both physics-informed neural networks and learnable optimization. The main Section~\ref{sec:LearnableOptimization} introduces the class of learnable optimizers used in this work. Section~\ref{sec:ResultsLearnableOptimization} contains the numerical results obtained by using these meta-trained optimizers for solving a variety of differential equations using physics-informed neural networks. A summary with a discussion on further possible research directions can be found in the final Section~\ref{sec:ConclusionsLearnableOptimization}.

\section{Solving differential equations with neural networks}\label{sec:PINNsOptimization}

The numerical solution of differential equations with neural networks was first proposed in~\cite{laga98a}. In this algorithm, the trial solution is brought into a form that accounts for initial and/or boundary conditions (as hard constraints), with the actual solution being found upon minimizing the mean-squared error that is defined as the residual of the given differential equations evaluated over a finite number of collocation points which are distributed over the domain of the problem. This method was recently popularized by~\cite{rais19a}, coining the term \textit{physics-informed neural networks}, and extended to also allow for the identification of differential equations from data. A recent review on this subject can be found in~\cite{cuom22a}.

More formally, consider the following initial--boundary value problem for a general system of $L$ partial differential equations of order $n$,
\begin{align}\label{eq:GeneralSystemOfEquations}
\begin{split}
& \Delta^l(t,\mathbf{x},\mathbf{u}_{(n)}) = 0, \qquad l=1,\dots, L,\quad t\in[0,t_{\rm f}],\ \mathbf{x}\in\Omega,\\
& \mathsf{I}^{l_{\rm i}}(\mathbf{x},\mathbf{u}_{(n_{\rm i})|_{t=0}}) = 0,\qquad l_{\rm i} = 1,\dots,L_{\rm i},\quad \mathbf{x}\in\Omega,\\
&\mathsf{B}^{l_{\rm b}}(t,\mathbf{x},\mathbf{u}_{(n_{\rm b})})=0,\quad l_{\rm b}=1,\dots, L_{\rm b},\qquad t\in[0,t_{\rm f}],\ \mathbf{x}\in\partial\Omega,
\end{split}
\end{align}
where $t\in[0,t_{\rm f}]$  is the time variable, $\mathbf{x}=(x_1,\dots,x_d)\in\Omega$ is the tuple of spatial independent variables, $\mathbf{u}=(u^1,\dots, u^q)$ is the tuple of dependent variables, and $\mathbf{u}_{(n)}$ is the tuple of all derivatives of the dependent variables with respect to the independent variables of order not greater than $n$. The initial value operator is denoted by $\mathsf{I}=(\mathsf{I}^1,\dots\mathsf{I}^{L_{\rm i}})$ and $\mathsf{B}=(\mathsf{B}^1,\dots,\mathsf{B}^{L_{\rm b}})$ denotes the boundary value operator. The spatial domain is $\Omega$ and the final time is $t_{\rm f}$.

In the following, we consider evolution equations for which the initial value operator reduces to
\[
\mathsf{I} = \mathbf{u}(0,\mathbf{x}) - \mathbf{f}(\mathbf{x}),
\]
where $\mathbf{f}(\mathbf{x})=(f^1(\mathbf{x}),\dots, f^q(\mathbf{x}))$ is a fixed vector-valued function. We also consider Dirichlet boundary conditions of the form
\[
\mathsf{B} = \mathbf{u}(t,\mathbf{x}) - \mathbf{g}(t,\mathbf{x}),
\]
where $\mathbf{g}(t,\mathbf{x})=(g^1(\mathbf{t,x}),\dots, g^q(t,\mathbf{x}))$ is another fixed vector-valued function.

Solving system~\eqref{eq:GeneralSystemOfEquations} with a neural network $\mathcal N^{\boldsymbol\theta}$ requires the parameterization of the solution of this system in the form $\mathbf{u}^{\boldsymbol \theta} = \mathcal N^{\boldsymbol \theta}(t,\mathbf{x})$, where the weights $\boldsymbol{\theta}$ of the neural network are found upon minimizing the loss function
\begin{subequations}\label{eq:LossFunction}
\begin{equation}\label{eq:LossFunctionCompositeLoss}
\mathcal L(\boldsymbol{\theta}) = \mathcal L_\Delta(\boldsymbol{\theta}) + \gamma_{\rm i}\mathcal L_{\rm i}(\boldsymbol{\theta}) + \gamma_{\rm b}\mathcal L_{\rm b}(\boldsymbol{\theta}).
\end{equation}
Here
\begin{align}\label{eq:LossFunctionContributions}
\begin{split}
\mathcal L_\Delta(\boldsymbol{\theta}) &= \frac{1}{N_\Delta}\sum_{i=1}^{N_\Delta}\sum_{l=1}^L\big|\Delta^l\big(t^i_\Delta,\mathbf{x}^i_\Delta,\mathbf{u}^{\boldsymbol{\theta}}_{(n)}(t^i_\Delta,\mathbf{x}^i_\Delta)\big)\big|^2,\\
\mathcal L_{\rm i}(\boldsymbol{\theta}) &= \frac{1}{N_{\rm i}}\sum_{i=1}^{N_{\rm i}}\sum_{l_{\rm i}=1}^{L_{\rm i}}\big|\mathsf{I}^{l_{\rm i}}\big(\mathbf{x}^i_{\rm i},\mathbf{u}_{(n_{\rm i})}^{\boldsymbol{\theta}}(0,\mathbf{x}^i_{\rm i})\big)\big|^2,\\
\mathcal L_{\rm b}(\boldsymbol{\theta}) &= \frac{1}{N_{\rm b}}\sum_{i=1}^{N_{\rm b}}\sum_{l_{\rm b}=1}^{L_{\rm b}}\big|\mathsf{B}^{l_{\rm i}}\big(t^i_{\rm b},\mathbf{x}^i_{\rm b},\mathbf{u}_{(n_{\rm b})}^{\boldsymbol{\theta}}(t^i_{\rm b},\mathbf{x}^i_{\rm b})\big)\big|^2,
\end{split}
\end{align}
\end{subequations}
are the mean squared error losses corresponding to the differential equation, the initial condition  and the boundary value residuals, respectively, and $\gamma_{\rm i}$ and $\gamma_{\rm b}$ are positive scaling constants. These losses are evaluated over the collection of collocation points  $\big\{(t_\Delta^i,\mathbf{x}_\Delta^i)\big\}_{i=1}^{N_\Delta}$ for the system $\Delta$, $\big\{(0,\mathbf{x}_{\rm i}^i)\big\}_{i=1}^{N_{\rm i}}$ for the initial data, and $\big\{(t_{\rm b}^i,\mathbf{x}_{\rm b}^i)\big\}_{i=1}^{N_{\rm b}}$ for the boundary data, respectively. Upon successful minimization, the neural network $\mathcal N^{\boldsymbol\theta}$ provides a numerical parameterization of the solution of the given initial--boundary value problem.

\section{Related work}\label{sec:RelatedWorkOptimization}

Physics-informed neural networks were proposed in~\cite{laga98a}, and popularized through the work~\cite{rais19a}, and have since been used extensively for solving differential equations in science and engineering. While the general algorithm for training neural networks to solve differential equations is straightforward, several complications arise in practice. Firstly, balancing the individual loss contributions in~\eqref{eq:LossFunctionContributions} so that all the initial values, the boundary values, and the differential equations are adequately enforced simultaneously constitutes a multi-task learning problem which may not be properly solved by minimizing the composite loss function~\eqref{eq:LossFunctionCompositeLoss}, see~\cite{sene18a,yu20a} for some work on multi-task learning problems. Secondly, it is well-known that training neural networks using gradient descent methods leads to a spectral bias in the form of low frequencies being learned first and high-frequencies requiring longer training times~\cite{raha19a}. Correspondingly, oscillatory solutions or stiff problems may not be accurately learned using standard physics-informed neural networks. Lastly, the general setup~\eqref{eq:LossFunction} requires proportionally more collocation points the larger the spatio-temporal domain of the differential equation being solved is. Training neural networks for solving differential equations over large spatio-temporal domains can destabilize training, which is frequently encountered in practice. In most cases, the solution for such problems is a trivial constant solution of the given differential equation~\cite{bihl22a,penw23a,wang22a}. One straightforward solution for this problem is to break the entire domain into multiple sub-domains, and solve a sequence of smaller problems with multiple neural networks instead. This multi-model approach has recently been used for solving the shallow-water equations on a rotating sphere~\cite{bihl22a}.

Learnable optimization has been the topic of research since the works~\cite{beng90a, beng95a}, with~\cite{andr16a} popularizing the use of neural network based \textit{learning to learn} optimization. The latter paper specifically introduced an LSTM-type neural network optimizer that is being trained using gradient descent. Subsequent work focussed on improving the performance of learnable neural network based optimizers by improving their training strategies, see e.g.~\cite{lv17a, vico21a}, improving the LSTM architecture of the optimizer~\cite{wich17a}, or replacing the LSTM-based architecture in favour of a simpler MLP-based one~\cite{harr22a, metz22a}. Below, we will use the optimizer proposed in~\cite{harr22a}. For a more comprehensive review on learnable optimization consult the recent review paper~\cite{chen22a}.

To the best of our knowledge, the use of learnable optimization for physics-informed neural networks has not been pursued so far. The related field of using meta-learning to accelerating the training of physics-informed neural networks has been investigated in~\cite{liu22a} and~\cite{psar22a} recently. Specifically, in these works the authors used meta-learning to discover suitable initialization methods and physics-informed neural network loss functions that generalize across relevant task distributions, respectively, thereby speeding up training of individual physics-informed neural networks from these task distributions.

\section{Meta-learnable optimization for physics-informed neural networks}\label{sec:LearnableOptimization}

A main goal of meta-learned optimization it to improve hand-designed optimization rules such as the Adam optimizer~\cite{king14a} for updating the weight vector $\boldsymbol{\theta}$ of a neural network with loss function~$L(\boldsymbol{\theta})$. Recall that the Adam update rule is given by
\begin{align*}
&\mathbf{m}_{t} = \beta_1 \mathbf{m}_{t-1} + (1-\beta_1)\nabla_{\boldsymbol{\theta}}L(\boldsymbol{\theta}_{t-1}),\quad \mathbf{v}_{t} = \beta_2 \mathbf{v}_{t-1} + (1-\beta_2)(\nabla_{\boldsymbol{\theta}}L(\boldsymbol{\theta}_{t-1}))^2,\\
&\mathbf{\hat m}_{t} = \mathbf{m}_{t}/(1-\beta_1^t),\quad \mathbf{\hat v}_{t} = \mathbf{v}_{t}/(1-\beta_2^t),\\ 
&\boldsymbol{\theta}_t = \boldsymbol{\theta}_{t-1} - \eta \mathbf{w}_{\rm adam} = \boldsymbol{\theta}_t - \eta \mathbf{\hat m}_{t}/(\sqrt{\mathbf{\hat v}_{t}} + \varepsilon),
\end{align*}
where $t=1,\dots,$ is the optimization time step, $\mathbf{m}$ and $\mathbf{v}$ are the first and second moment vectors, with $\beta_1,\beta_2\in[0,1)$ being the exponential decay rates for the moment estimates, $\varepsilon$ being a regularization constant, and $\eta$ being the learning rate.

Similarly, the parameter updates of a meta-learned optimizer is structured as
\begin{equation}\label{eq:LearnableOptimizerVector}
\boldsymbol{\theta}_t = \boldsymbol{\theta}_{t-1} - \mathbf{f}(\mathbf{z}_t;\boldsymbol{\vartheta}),
\end{equation}
where $\mathbf{f}$ is the parametric update function with $\mathbf{z}_t$ referring to the input features of the learnable optimizer, and $\boldsymbol{\vartheta}$ are the trainable meta-parameters of the optimizer, usually the weights of a neural network. To allow for the learnable optimizer to be transferable to neural networks of different sizes it is customary to have the parameter update rule~\eqref{eq:LearnableOptimizerVector} act component-wise, with each weight $\theta_i$ of the weight vector $\boldsymbol{\theta}$ being updated in the same way. Thus, in the following we describe the parameteric update formula in terms of scalar variables, rather than vector variables.

While there are several optimizer architectures that have been proposed in the literature~\cite{chen22a}, here we use a relatively simple multi-layer perceptron for the optimizer architecture. Notably, we follow the work~\cite{harr22a} and structure the parametric update formula for each weight $\theta_i$ as
\begin{equation}\label{eq:StarOptimizer}
f = \lambda_1\exp(\lambda_2 s^{\rm adam}_{\boldsymbol\vartheta})) w_{\rm adam} + \frac{\lambda_3}{\sqrt{v_t}+\varepsilon}d^{\rm bb}_{\boldsymbol\vartheta}\exp(\lambda_4 s^{\rm bb}_{\boldsymbol\vartheta}),
\end{equation}
where $\lambda_i$, $i=1,\dots,4$ are positive constants, $w_{\rm adam}$ corresponds to the Adam update step and $s^{\rm adam}_{\boldsymbol\vartheta}$, $s^{\rm bb}_{\boldsymbol\vartheta}$ and $d^{\rm bb}_{\boldsymbol\vartheta}$ are to the output heads of the meta-learned optimizer with neural network weights $\boldsymbol{\vartheta}$. 

On a high level, the first term in the learnable update formula~\eqref{eq:StarOptimizer} can be seen as a nominal term derived from the Adam update formula with scalable learning rate $\lambda_1\exp(\lambda_2 s^{\rm adam}_{\boldsymbol\vartheta}))$, which guarantees an update step in a descent direction, and the second term corresponds to a blackbox update term structured as the product of a directional and magnitudinal term, $d^{\rm bb}_{\boldsymbol\vartheta}$ and $\exp(\lambda_4 s^{\rm bb}_{\boldsymbol\vartheta})$, respectively, with the denominator $\sqrt{v_t}+\varepsilon$ acting as a preconditioner that should guarantee that the overall update formula leads corresponds to a descending on the loss surface. For more details on the rationale behind the update rule~\eqref{eq:StarOptimizer}, consult~\cite{harr22a}.
 
The inputs $\mathbf{z}_t$ at optimization step $t$ to the multi-layer perceptron optimizer with output heads $s^{\rm adam}_{\boldsymbol\vartheta}$, $s^{\rm bb}_{\boldsymbol\vartheta}$ and $d^{\rm bb}_{\boldsymbol\vartheta}$ are chosen as follows:
\begin{enumerate}\itemsep=0ex
\item The weights $\boldsymbol{\theta}_t$;
\item The gradients $\nabla_{\boldsymbol{\theta}}L(\boldsymbol{\theta}_t)$;
\item The second momentum accumulators $\mathbf{v}_t$ with decay rates $\beta_2\in\{0.5, 0.9, 0.99, 0.999\}$;
\item One over the square root of the above four second momentum accumulators;
\item The time step $t$.
\end{enumerate}
Here, we build upon the extensive study carried out in~\cite{metz22a}, with the above input parameters heuristically being found to perform well for the physics-informed neural networks that were trained in this work.

All input features (except the time step) were normalized to have a second moment of one. The time step is converted into a total of 11 features by computing $\tanh(t/x)$ where $x\in\{1,3,10,30,100,300,1000,3000,10k,30k,100k\}$. All features were then concatenated and passed through a standard multi-layer perceptron to yield the above three output heads.

\section{Numerical results}\label{sec:ResultsLearnableOptimization}

In this section we showcase the use of meta-learned optimization for solving some well-known differential equations from mathematical physics, that have been extensively studied using physics-informed neural networks. In all of the following examples we use the vanilla version of physics-informed neural networks as laid out in~\cite{laga98a,rais19a}. As discussed in Section~\ref{sec:RelatedWorkOptimization}, it is well-understood by now that this formulation can suffer from several drawbacks which to remedy is currently an active research field. As such, the goal of this section is not to obtain the best possible numerical solution for each given model, but to show how meta-learned optimization can improve the results obtainable using vanilla physics-informed neural network when compared to using standard optimization. Our base optimizer we compare against is the Adam optimizer, the de-facto standard being used in the field of physics-informed neural networks today. 

In all examples below, the output heads of the meta-learned optimizer were initialized using a normal distribution with zero mean and variance of $10^{-3}$, to guarantee that the neural network output is close to zero at the beginning of meta-training of the optimizer. Due to the form of the meta-learned optimizer~\eqref{eq:StarOptimizer}, this means that before meta-training starts, the meta-learned optimizer is very close to the standard Adam optimizer. 

For all examples, the multi-layer perceptron being used for the meta-learned optimizer has two hidden layers with 32 units each, using the swish activation function. This architecture was found using hyperparameter tuning to give a good balance between computational overhead of meta-training the optimizer and error level of the resulting optimizer. We should like to note here that in contrast to the application of meta-learned optimization in areas of modern deep learning, such as computer vision or natural language processing, which work with neural networks with up to hundreds of hidden layers and billions of weights, the neural networks arising in physics-informed neural networks are typically relatively small. In fact, all of the architectures considered in this paper have less than 10,000 trainable parameters. This allows for larger neural networks being used for the meta-learned optimizer, without incurring computationally infeasible costs. Still, the underlying multi-layer perceptron of the meta-learned optimizer is relatively small, having only 2,115 trainable parameters.

We train this optimizer using the \textit{persistent evolutionary strategy}, a zeroth-order stochastic optimization method described in~\cite{vico21a}. This algorithm has several hyperparameters, including the total number of particles $N$ used for gradient computation, the partial unroll length $K$ of the inner optimization problem before a meta-gradient update is computed, the standard deviation of perturbations $\sigma$ and the learning rate $\alpha$ for the meta-learned weight update. Using hyperparamter tuning, we determined $N=2$ (antithetic) particles, $K=1$ epochs and a learning rate of $\alpha=10^{-4}$ to be the best hyperparameters for our problem. For more details, see Algorithm~2 in~\cite{vico21a}. 

For each problem, unless otherwise specified, we then sample a total of 20 different tasks and train the meta-learned optimizer for a total of 50 epochs on the associated tasks. Each task corresponds to a new instantiation of the particular neural network architecture for the same equation parameters, meaning the only difference in each task are the initial (random) weights of the neural network model. We found empirically that training the meta-learned optimizer for relatively few epochs (50 epochs compared to using the learned optimizer for more than 1000 epochs at testing stage) provided a good balance between performance and meta-training cost. To guarantee a fair comparison, at testing time the initial weights of the two neural networks being trained with the respective optimizers are exactly the same. 

In Table~\ref{tab:ArchitecturesOptimization} we summarize the parameters of the physics-informed neural networks trained in this section. We use hyperbolic tangents as activation function for all hidden layers. We use mini-batch gradient computation with a total of 10 batches per epoch.

\begin{table}[!ht]
\centering
\caption{Parameters of the physics-informed neural networks trained below.}
\begin{tabular}{l|cccc}
& Linear advection Eq. & Poisson Eq. & KdV Eq. & Burgers Eq. \\
\hline
\hline
\# hidden layers & 2 & 4 & 6 & 6 \\ 
\hline
\# units & 20 & 20 & 20 & 20 \\
\hline
\# PDE points & 10,000 & 10,000 & 10,000 & 10,000 \\
\hline
\# IC/BC points & 100 & 400 & 100 & 100 \\
\hline
\# epochs & 3,000 & 2,000 & 1,000 & 1,000 \\
\hline
\end{tabular}
\label{tab:ArchitecturesOptimization}
\end{table}

We report both the time series of the loss for the standard Adam optimizer and the meta-learned optimizer, and the error $e=u_{\rm nn} - u_{\rm ref}$, where $u_{\rm ref}$ is either the analytical solution (if available), or a high-resolution numerical reference solution obtained from using a pseudo-spectral method for the spatial discretization and an adaptive Runge--Kutta method for time stepping using the method of lines approach~\cite{durr10a}.

The algorithm described here has been implemented using \texttt{TensorFlow} 2.11 and the codes will be made available on GitHub\footnote{\url{https://github.com/abihlo/LearnableOptimizationPinns}}.

\subsection{One-dimensional linear advection equation}

As a first example, consider the one-dimensional linear advection equation
\[
u_t + cu_x = 0,
\]
where we consider $t\in[0,3]$ and $x\in[-1,1]$ with $c=1$ being the advection velocity. We set $u(0,x)=u_0(x)=\cos\pi x$ and use periodic boundary conditions. We enforce the periodic boundary conditions as hard constraint in the physics-informed neural networks, using the strategy introduced in~\cite{bihl22a}. We set $\gamma_{\rm i}=1$ in the loss function~\eqref{eq:LossFunctionCompositeLoss}. The learning rate for Adam was $\eta=10^{-3}$. The constants of the learnable optimizer were all chosen as $\lambda_i=10^{-3}$, $i=1,\dots,4$.

\begin{figure}[!ht]
\centering
\includegraphics[scale=0.5]{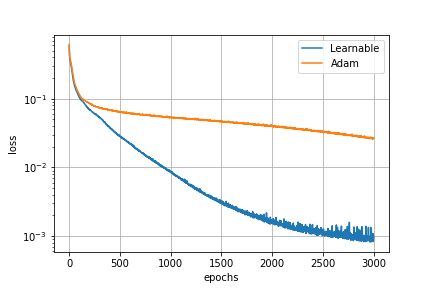}
\caption{Training loss for the Adam and meta-learned optimizers for the linear advection equation.}
\label{fig:TrainingLossLinearAdvectionEquation}
\end{figure}

\begin{figure}[!ht]
\centering
\includegraphics[scale=0.3]{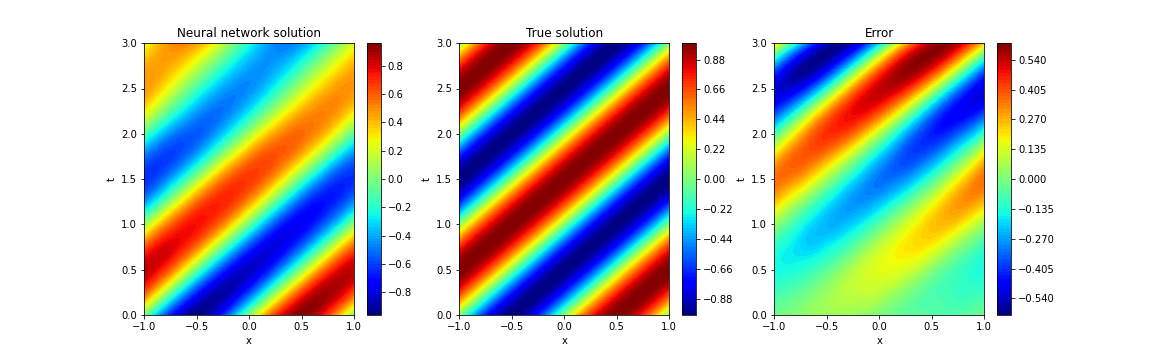}
\includegraphics[scale=0.3]{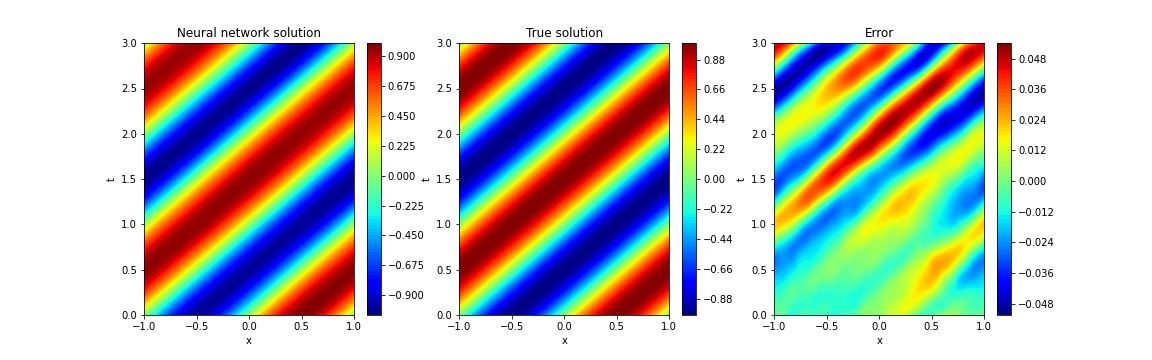}
\caption{Numerical results for the linear advection equation. \textit{Top row:} Standard Adam optimizer. \textit{Bottom row:} Meta-learned optimizer. Left to right shows the numerical solution obtained from the physics-informed neural networks, the exact solution, and the difference between the numerical solution and the exact solution.}
\label{fig:NumericalResultsLinearAdvectionEquation}
\end{figure}

The numerical results for this example are depicted in Figures~\ref{fig:TrainingLossLinearAdvectionEquation} and~\ref{fig:NumericalResultsLinearAdvectionEquation}. For this particular example, the meta-learned optimizer considerably outperforms the standard Adam optimizer, resulting in a training loss and point-wise error that is more than 10 times smaller. The loss for the physics-informed neural network using the meta-learned optimizer after 500 epochs is lower than the final loss after 3000 epochs for the respective network using Adam.

\subsection{Poisson equation}

As an example for a boundary-value problem, consider the two-dimensional Poisson equation
\[
u_{xx} + u_{yy} = f(x,y),
\]
over the domain $\Omega=[-1,1]\times[-1,1]$ for the exact solution
\[
u_{\rm exact}(x,y) = (0.1\sin 2\pi x + \tanh 10 x)\sin2\pi y,
\]
with the associated right-hand side using Dirichlet boundary conditions. This problem was considered in~\cite{khar21a}. Since this is a boundary value problem, there is no initial loss in the loss function~\eqref{eq:LossFunctionCompositeLoss} and we use $\gamma_{\rm b}= 1000$. This value was chosen heuristically to balance the differential equation and boundary value losses. The learning rate for Adam for this example was set to $\eta=10^{-3}$ and so were the constants of the meta-learned optimizer, $\lambda_i=10^{-3}$, $i=1,\dots,4$.

\begin{figure}[!ht]
\centering
\includegraphics[scale=0.5]{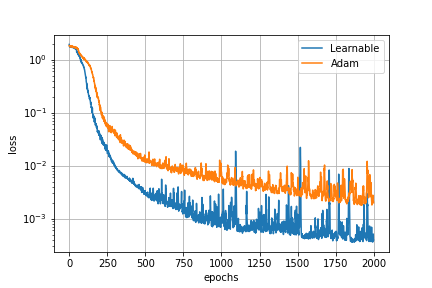}
\caption{Training loss for the Adam and meta-learned optimizers for the two-dimensional Poisson equation.}
\label{fig:TrainingLossPoissonEquation}
\end{figure}

\begin{figure}[!ht]
\centering
\includegraphics[scale=0.3]{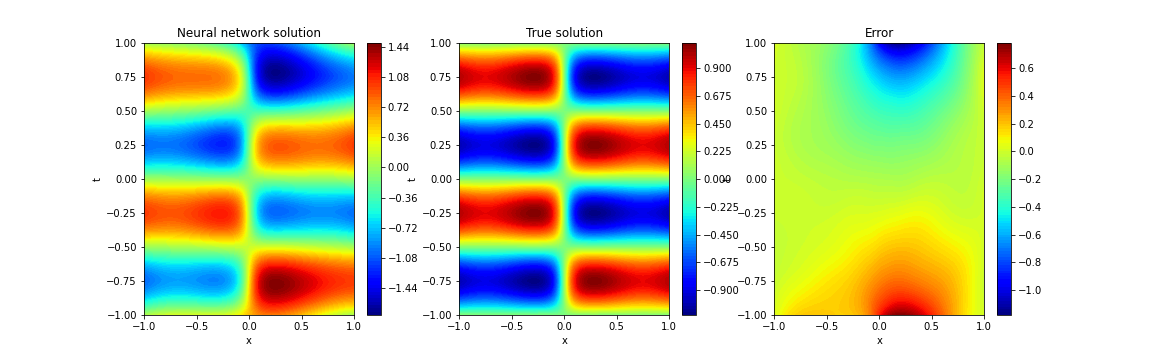}
\includegraphics[scale=0.3]{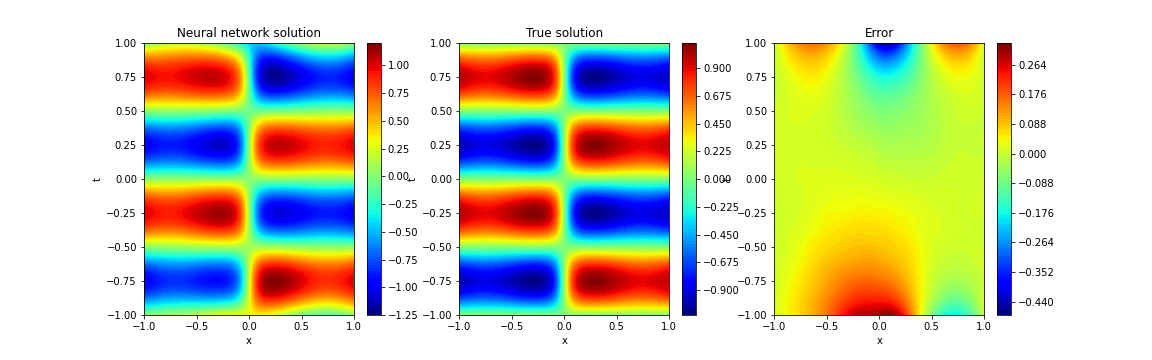}
\caption{Numerical results for the Poisson equation. \textit{Top row:} Standard Adam optimizer. \textit{Bottom row:} Meta-learned optimizer. Left to right shows the numerical solution obtained from the physics-informed neural networks, the exact solution, and the difference between the numerical solution and the exact solution.}
\label{fig:NumericalResultsPoissonEquation}
\end{figure}

The training loss for this example is shown in Fig.~\ref{fig:TrainingLossPoissonEquation}, the numerical results as compared to the exact solution with the associated point-wise error are depicted in Fig.~\ref{fig:NumericalResultsPoissonEquation}. As with the linear advection equation from the previous example, also for the Poisson equation the meta-learned optimization method leads to better results both in terms of a lower training loss and smaller point-wise errors compared to the standard Adam optimizer.

\subsection{Korteweg--de Vries equation}

We next consider the Korteweg--de Vries equation
\[
u_t + uu_x - \nu u_{xxx} = 0,
\]
with initial condition $u(0,x) = -\sin\pi x$ using periodic boundary conditions over the domain $x\in[-1,1]$ and $t\in[0,1]$, setting $\nu=0.0025$. This equation has been extensively studied using physics-informed neural networks, see e.g.~\cite{jagt20a,rais19a}. Again, we enforce the periodic boundary conditions as hard constraint and set $\gamma_{\rm i}=1$ in the loss function~\eqref{eq:LossFunctionCompositeLoss}. The learning rate of the Adam optimizer was chosen as $\eta=5\cdot 10^{-4}$, and the constants of the meta-learned optimizer were set to $\lambda_1 = 5\cdot 10^{-4}$ and $\lambda_i=10^{-3}$, $i=2,\dots,4$.

\begin{figure}[!ht]
\centering
\includegraphics[scale=0.5]{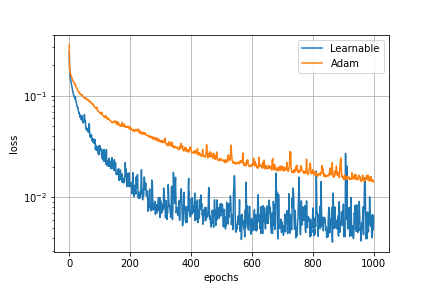}
\caption{Training loss for the Adam and meta-learned optimizers for the Korteweg--de Vries equation.}
\label{fig:TrainingLossKdVEquation}
\end{figure}

\begin{figure}[!ht]
\centering
\includegraphics[scale=0.3]{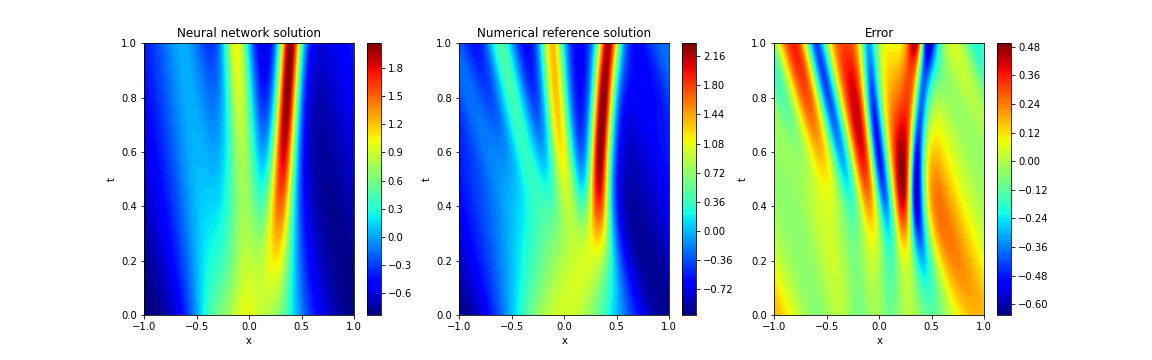}
\includegraphics[scale=0.3]{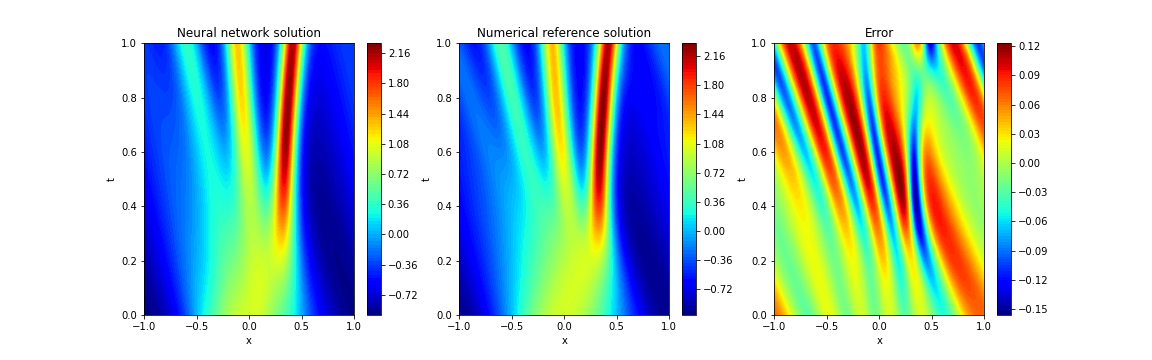}
\caption{Numerical results for the Kortweg--de Vries equation. \textit{Top row:} Standard Adam optimizer. \textit{Bottom row:} Meta-learned optimizer. Left to right shows the numerical solution obtained from the physics-informed neural networks, the numerical reference solution, and the difference between the numerical solution and the reference solution.}
\label{fig:NumericalResultsKdVEquation}
\end{figure}

Figure~\ref{fig:TrainingLossKdVEquation} contains the respective training losses of the Adam and meta-learned optimizers. The numerical solutions for the associated trained physics-informed neural networks as compared against the numerical solution obtained from a pseudo-spectral numerical integration method are featured in Figure~\ref{fig:NumericalResultsKdVEquation}. These plots again illustrate that the meta-learned optimizer reduces the training loss considerably faster than the standard Adam optimizer, which also improves upon the point-wise error of the numerical solution compared to the reference solution. In fact, the training loss after 200 epochs is lower for the meta-learned optimizer than what the Adam optimizer achieves at the end of training.

We next consider a problem of transfer learning for meta-learned optimizers for the Korteweg--de Vries equation. A common task in the numerical solution of differential equation is to change the initial condition of the problem. For physics-informed neural networks this requires re-training of the network, which is computationally costly. To investigate this problem, we sample our task distribution for meta-training the optimizer from an ensemble of initial conditions here. For the sake of simplicity we consider initial conditions of the form
\[
u(0,x) = \cos (kx+\phi),
\]
where $k$ is sampled from integers between 1 and 3 and $\phi$ is sampled uniformly from $[-\pi/2,\pi/2]$. We choose a relatively narrow task distribution to speed up meta-learning. Once trained, we evaluate the optimizer on the unseen test problem with $k=2$ and $\phi=-\pi/4$. Since this is a harder problem than using the meta-learned optimizer on the same problem (i.e.\ same initial condition and same differential equation), we meta-train the optimizer on a total of 75 tasks here instead of the 20 tasks used so far.

\begin{figure}[!ht]
\centering
\includegraphics[scale=0.5]{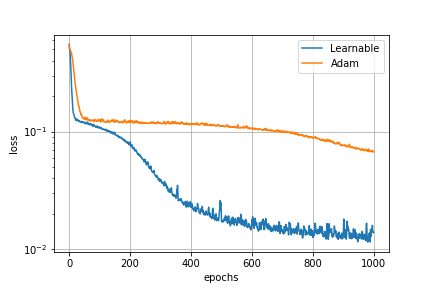}
\caption{Training loss for the Adam and meta-learned optimizers for the Korteweg--de Vries equation using transfer learning.}
\label{fig:TrainingLossKdVEquationTransfer}
\end{figure}

\begin{figure}[!ht]
\centering
\includegraphics[scale=0.3]{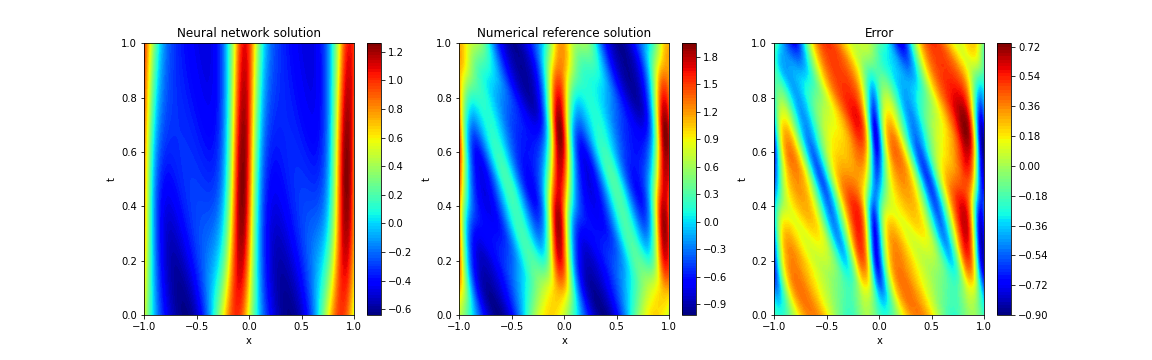}
\includegraphics[scale=0.3]{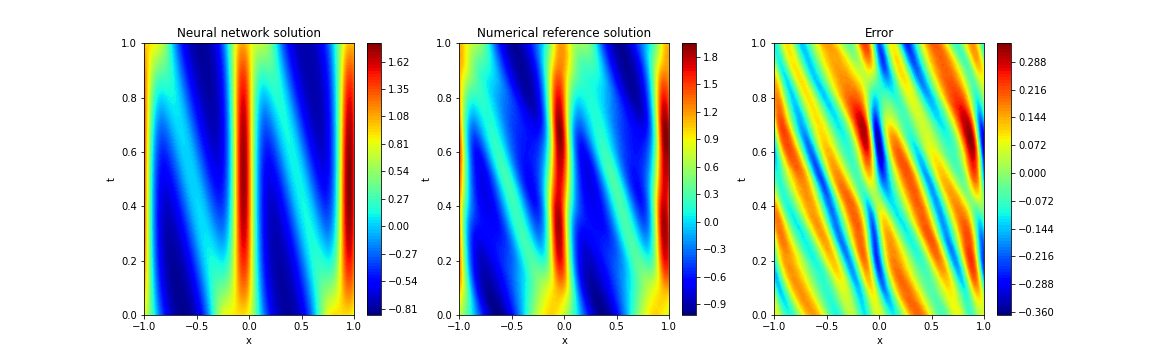}
\caption{Numerical results for the Kortweg--de Vries equation using transfer learning. \textit{Top row:} Standard Adam optimizer. \textit{Bottom row:} Meta-learned optimizer. Left to right shows the numerical solution obtained from the physics-informed neural networks, the numerical reference solution, and the difference between the numerical solution and the reference solution.}
\label{fig:NumericalResultsKdVEquationTransfer}
\end{figure}

The results of this experiment are depicted in Figures~\ref{fig:TrainingLossKdVEquationTransfer} and~\ref{fig:NumericalResultsKdVEquationTransfer}. These figures again show improvement of the meta-learned optimizer when compared to the results obtained using Adam. This demonstrates that transfer learning across the same equation class, i.e.\ choosing different initial values but keeping the equation the same, is indeed feasible. Moreover, the loss level achieved after 200 epochs using the meta-learned optimizer is comparable to the loss level obtained using the Adam optimizer after 1000 epochs, again pointing to the possibility of significant speed-up in training physics-informed neural networks. In the next example we show that transfer learning across different equation classes is possible as well.

\subsection{Burgers' equation}

As a last example we consider Burgers' equation
\[
u_t + uu_x -\nu u_{xx} = 0,
\]
over the temporal-spatial domain $[0,1]\times[-1,1]$ with initial condition $u(0,x) = -\sin\pi x$ and periodic boundary conditions in $x$-direction. The diffusion parameter was set as $\nu=0.01/\pi$. Burgers equation is also one of the most prominent examples considered using physics-informed neural networks, see~\cite{rais19a} for some results.

As for the Korteweg--de Vries equation, we enforce the periodic boundary conditions as hard constraints, use $\gamma_{\rm i}=1$ in the loss function~\eqref{eq:LossFunctionCompositeLoss}, and set the learning rate of the Adam optimizer to $\eta=5\cdot 10^{-4}$, and the constants of the meta-learned optimizer to $\lambda_1 = 5\cdot 10^{-4}$ and $\lambda_i=10^{-3}$, $i=2,\dots,4$.

Here we consider two meta-learned optimizers. The first is being trained as for the previous example, i.e.\ using Burgers' equation on 20 tasks, which each task being a newly instantiated neural network with different random initial weights. The second one is being meta-trained using the linear advection equation. This second optimizer should assess the transfer learning abilities of meta-trained optimizers across different differential equations. For this optimizer, we choose our tasks for varying advection velocities sampled uniformly from $c\in[-1,1]$. At testing time, this optimizer meta-trained on the linear advection equation is also used to train a physics-informed neural network for Burgers' equation.

\begin{figure}[!ht]
\centering
\includegraphics[scale=0.5]{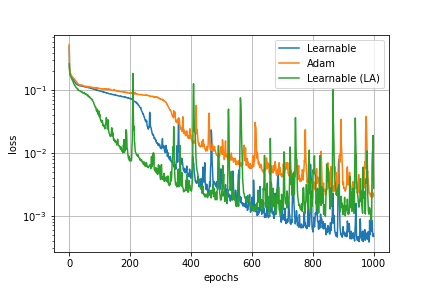}
\caption{Training loss for the Adam and meta-learned optimizers. We train two meta-learned optimizers for this case, one using the linear advection equation (green curve) to assess the transfer learning abilities of meta-learned optimizers, and one using Burgers equation itself (blue curve).}
\label{fig:TrainingLossBurgersEquation}
\end{figure}

\begin{figure}[!ht]
\centering
\includegraphics[scale=0.3]{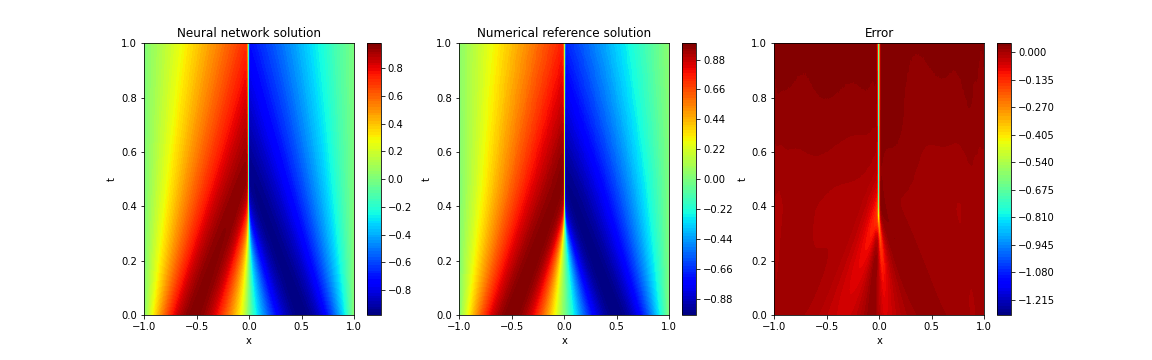}
\includegraphics[scale=0.3]{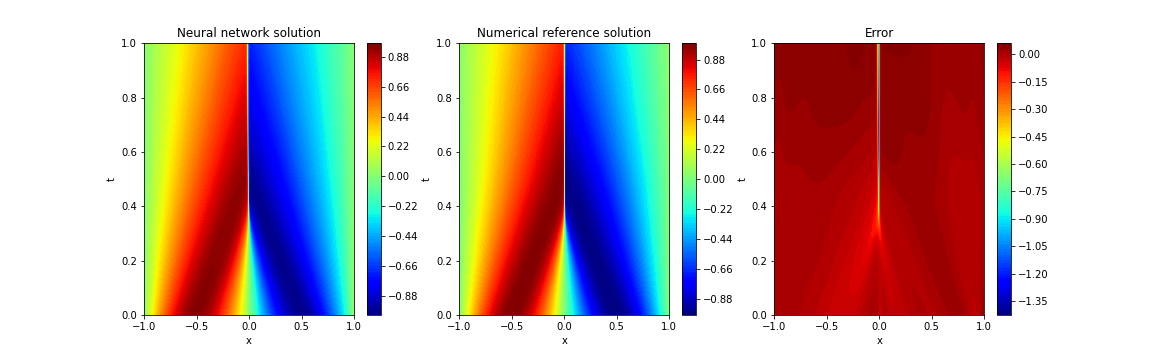}
\includegraphics[scale=0.3]{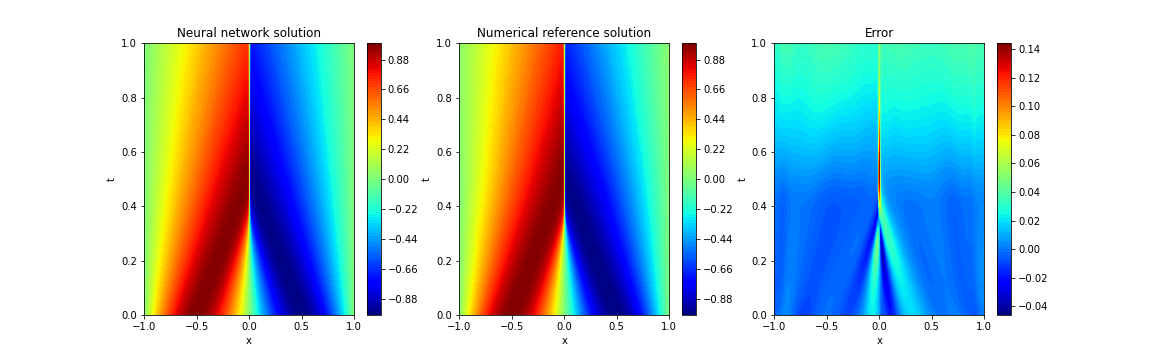}
\caption{Numerical results for Burgers' equation. \textit{Top row:} Standard Adam optimizer.
\textit{Middle row:} Meta-learned optimizer using the linear advection equation. \textit{Bottom row:} Meta-learned optimizer using Burgers equation. Left to right shows the numerical solution obtained from the physics-informed neural networks, a numerical reference solution, and the difference between the numerical solution and the reference solution.}
\label{fig:NumericalResultsBurgersEquation}
\end{figure}

Figures~\ref{fig:TrainingLossBurgersEquation} and~\ref{fig:NumericalResultsBurgersEquation} contain the associated numerical results for this example, showing the training loss of the three respective optimizers and the actual numerical results for solving Burgers' equation using the trained neural networks. Figures~\ref{fig:TrainingLossBurgersEquation} illustrates that the meta-learned optimizer trained using the linear advection equation also outperforms the Adam optimizer. Interestingly, this optimizer also outperforms the meta-learned optimizer trained on Burgers' equation over the first 400 epochs, although exhibiting substantially higher oscillations than the latter. At the end of the training, the loss for the linear advection trained meta-learned optimizer is still a bit lower than Adam, although the loss seems to have stagnated after about 600 epochs of training. 

The loss levels are also consistent with the numerical results shown in Fig.~\ref{fig:NumericalResultsBurgersEquation}, illustrating that the meta-learned optimizer using Burgers' equation is the best with the other two optimizers yielding comparable errors. Still, these results show the transfer learning abilities of meta-learned optimizers across different differential equations, which could be leveraged in a multitude of ways. For the particular example of Burgers' equation, the meta-learned optimizer trained on the linear advection equation could be used for the first few hundred epochs, before being chained with another optimizer more suitable for longer training. It is also conceivable that more extensive meta-training, either using more tasks sampled from a wider task distribution, or from wider classes of differential equations altogether, could give optimizers that are applicable to more than a single class of differential equations.

\section{Conclusion}\label{sec:ConclusionsLearnableOptimization}

We have investigated the use of meta-learned optimization for improving the training of physics-informed neural networks in this work. Meta-learned optimization, or learning to learn, has become an increasingly popular topic in deep learning and thus it is natural to investigate its applicability in scientific machine learning as well. We have done so here by illustrating that meta-learned optimization can be used to improve the numerical results obtainable using physics-informed neural networks, which is a popular machine learning-based method for solving differential equations. We have also provided proof-of-concept that these meta-learned optimizers have transfer learning capabilities, i.e.\ that they can be used for problems that are different from those they were trained on.

The goal of this paper was to illustrate that meta-learned optimization alone can substantially improve the vanilla form of physics-informed neural networks, which was laid out in the seminal works~\cite{laga98a,rais19a}. This form has been extensively studied, and we have shown here that meta-learned optimization can give (sometimes substantially) better numerical results compared to standard hand-crafted optimization rules. This means that meta-learned optimizers are able to reach a particular error level quicker than standard optimizers, resulting in either shorter training times (for a given target computational error) or better numerical accuracy (for the same number of training epochs). 

There are several avenues for future research that would provide natural extensions to the present work. Firstly, one could investigate the use of meta-learned optimization for other formulations of physics-informed neural networks. We have refrained from doing so here, as there is not one canonical formulation of improved training strategies for physics-informed neural networks but rather a zoo of methods that are applicable to different classes of differential equations. This list of methods includes, to name a few, variational formulations~\cite{khab09Ay}, formulations based on domain decompositions~\cite{jagt20a}, formulations based on improved loss functions~\cite{psar22a,wang22a}, re-sampling strategies~\cite{wu2023a}, and operator-based formulations~\cite{wang23a}. It should also be stressed that while these formulations can considerably outperform vanilla physics-informed networks, the latter are still extensively being used in the literature today, see~\cite{cuom22a} for a recent review. 

Secondly, there is a multitude of other meta-learned optimization algorithms based on neural networks that have been proposed in the literature, see the review paper~\cite{chen22a} for an extensive list of such optimizers. There are also several training strategies available for meta-learned optimization, including gradient descent, evolutionary strategies and reinforcement learning based ones~\cite{chen22a}. 

Together, this provides a rich set of training strategies, meta-learnable optimizer architectures and physics-informed model formulations that could be explored together to possibly find more accurate solutions of differential equations using physics-informed neural networks. We plan to explore some of these possibilities in the near future. 

\section*{Acknowledgements}

This research was undertaken thanks to funding from the Canada Research Chairs program and the NSERC Discovery Grant program.

{\footnotesize\setlength{\itemsep}{0ex}

}

\end{document}